\documentclass[letterpaper, 10 pt, conference]{ieeeconf}  % Comment this line out if you need a4paper
\pdfoutput=1
\IEEEoverridecommandlockouts                              % This command is only needed if 
\usepackage[utf8]{inputenc}
\usepackage[T1]{fontenc}

\usepackage[pdftex]{graphicx}
\usepackage{epstopdf}
\usepackage{nicefrac}
\usepackage{booktabs}
\usepackage{multirow}
\usepackage{siunitx}
\usepackage{microtype}
\usepackage[vlined,ruled]{algorithm2e}
\usepackage[table,xcdraw]{xcolor}
\usepackage{multicol}
\usepackage{fancyhdr}
\usepackage{verbatim} 
\usepackage{adjustbox}
\usepackage{balance}
\usepackage{amsmath}
\usepackage{amssymb}

\DeclareMathOperator*{\argmin}{\arg\!\min}
\DeclareMathOperator*{\argmax}{\arg\!\max}
\DeclareMathOperator{\diag}{\operatorname{diag}}

\DeclareMathOperator{\meanOp}{\operatorname{mean}}

\newcommand{\vect}[1]{\boldsymbol{\mathbf{#1}}}
\newcommand{\est}[1]{\hat{#1}}
\newcommand{\opt}[1]{#1^\ast}

\newcommand{\State}{\vect{X}}
\newcommand{\StateOpt}{\vect{\opt{X}}}
\newcommand{\StateEst}{\vect{\est{X}}}

\newcommand{\state}{\vect{x}}
\newcommand{\stateOf}[2]{\state^{#1}_{#2}}
\newcommand{\statei}{\stateOf{}{i}}
\newcommand{\stateti}{\stateOf{}{t,i}}

\newcommand{\statej}{\stateOf{}{j}}

\newcommand{\stateConf}{x^{conf}_{t,i}}
\newcommand{\stateDim}{\stateOf{dim}{t,i}}
\newcommand{\stateDimPast}{\stateOf{dim}{t-1,i}}

\newcommand{\statePos}{\stateOf{pos}{t,i}}
\newcommand{\statePosFut}{\stateOf{pos}{t+1,i}}
\newcommand{\statePosPast}{\stateOf{pos}{t-1,i}}
\newcommand{\statePosn}{\stateOf{pos}{t,n}}
\newcommand{\statePosm}{\stateOf{pos}{t,m}}
\newcommand{\stateVel}{\stateOf{vel}{t,i}}
\newcommand{\stateVelFut}{\stateOf{vel}{t+1,i}}

\newcommand{\Measurement}{\vect{Z}}

\newcommand{\measurement}{\vect{z}}
\newcommand{\measurementOf}[2]{\measurement^{#1}_{#2}}
\newcommand{\measurementi}{\measurementOf{}{i}}

\newcommand{\measurementtj}{\measurementOf{}{t,j}}
\newcommand{\measurementPos}{\measurementOf{pos}{t,j}}
\newcommand{\measurementDim}{\measurementOf{dim}{t,j}}
\newcommand{\measurementConf}{z^{conf}_{t,j}}
\newcommand{\measurementO}{z_0}
\newcommand{\error}{\vect{e}}
\newcommand{\errorOf}[2]{\error^{#1}_{#2}}
\newcommand{\errori}{\errorOf{}{i}}

\newcommand{\errorRep}{\errorOf{rep}{t,n,m}}
\newcommand{\errorDet}{\errorOf{det}{t,i}}
\newcommand{\errorCV}{\errorOf{cv}{t,i}}

\newcommand{\prob}{\vect{P}}
\newcommand{\probOf}[2]{\vect{P}\left( #1 | #2\right) }

\newcommand{\cov}{\vect{\Sigma}}
\newcommand{\covDetNull}{\vect{\Sigma}^{det}_{0}}
\newcommand{\covCV}{\vect{\Sigma}^{cv}}
\newcommand{\covRep}{\vect{\Sigma}^{rep}}
\newcommand{\covDet}{\vect{\Sigma}^{det}}
\newcommand{\mean}{{\vect{\mu}}}
\newcommand{\meanDetNull}{{\vect{\mu}}_{0}}
\newcommand{\weight}{w}
\newcommand{\info}{\vect{\mathcal{I}}}
\newcommand{\sqrtinfo}{\info^{\frac{1}{2}}}
\newcommand{\sqrtinfoOf}[1]{\info_{#1}^{\frac{1}{2}}}
\newcommand{\sqrtinfoj}{\sqrtinfoOf{j}}

\newcommand{\Params}{\vect{\theta}}
\newcommand{\ParamsSingle}{\theta_{i,j}}

\newcommand{\threshDist}{d_{min}}
\newcommand{\threshDet}{n_{det}}
\newcommand{\threshLost}{n_{lost}}
\newcommand{\threshPerm}{n_{perm}}
\newcommand{\threshConf}{c_{min}}

\newcommand{\bestResult}[1]{{\color[HTML]{33A151} \textbf{#1}}}

\makeatletter
\let\NAT@parse\undefined
\makeatother
\usepackage[numbers,sort,square]{natbib}

\PassOptionsToPackage{hyphens}{url}
\usepackage{hyperref}

\def\equationautorefname~#1\null{(#1)\null}

\title{\LARGE \bf Factor Graph based 3D Multi-Object Tracking in Point Clouds}

\author{\authorblockN{Johannes Pöschmann, Tim Pfeifer and Peter Protzel}%
\authorblockA{Dept.~of Electrical Engineering and Information Technology\\
TU Chemnitz, Germany\\
Email: \{firstname.lastname\}@etit.tu-chemnitz.de}}

\begin{document}

\maketitle
% copyright notice 
\thispagestyle{fancy}
\fancyhf{}
\fancyhead[OL]{ 
    \footnotesize
	To appear in Proc. of IEEE Intl. Conf. on Intelligent Robots and Systems (IROS), 2020, Las Vegas, USA.\\
	\tiny
	\copyright 2020 IEEE. Personal use of this material is permitted. Permission from IEEE must be obtained for all other uses, in any current or future media, including reprinting/republishing this material for advertising or promotional purposes, creating new collective works, for resale or redistribution to servers or lists, or reuse of any copyrighted component of this work in other works.}
\setlength{\headheight}{20pt}

%%%%%%%%%%%%%%%%%%%%%%%%%%%%%%%%%%%%%%%%%%%%%%%%%%%%%%%%%%%%%%%%%%%%%%%%%%%%%%%%
\begin{abstract}

Accurate and reliable tracking of multiple moving objects in 3D space is an essential component of urban scene understanding.
This is a challenging task because it requires the assignment of detections in the current frame to the predicted objects from the previous one.
Existing filter-based approaches tend to struggle if this initial assignment is not correct, which can happen easily.

We propose a novel optimization-based approach that does not rely on explicit and fixed assignments.
Instead, we represent the result of an off-the-shelf 3D object detector as Gaussian mixture model, which is incorporated in a factor graph framework.
This gives us the flexibility to assign all detections to all objects simultaneously.
As a result, the assignment problem is solved implicitly and jointly with the 3D spatial multi-object state estimation using non-linear least squares optimization.

Despite its simplicity, the proposed algorithm achieves robust and reliable tracking results and can be applied for offline as well as online tracking.
We demonstrate its performance on the real world KITTI tracking dataset and achieve better results than many state-of-the-art algorithms.
Especially the consistency of the estimated tracks is superior offline as well as online.

\end{abstract}

%%%%%%%%%%%%%%%%%%%%%%%%%%%%%%%%%%%%%%%%%%%%%%%%%%%%%%%%%%%%%%%%%%%%%%%%%%%%%%%%
\section{Introduction}
Robust tracking of multiple moving objects is a crucial component of urban scene understanding. 
Advances in autonomous driving are based on a reliable perception of dynamic objects around the ego-vehicle.
Therefore, fast and robust 3D online multi-object trackers (MOT) are required.
Another use case for 3D-MOT in urban scene understanding is the generation of reference data.
This requires an even more accurate and reliable tracking to support or replace labeling by humans, but can be applied offline.

\begin{figure}[htpb]
    \centering
    \vspace{0.3cm}
    \includegraphics[width=0.5\linewidth]{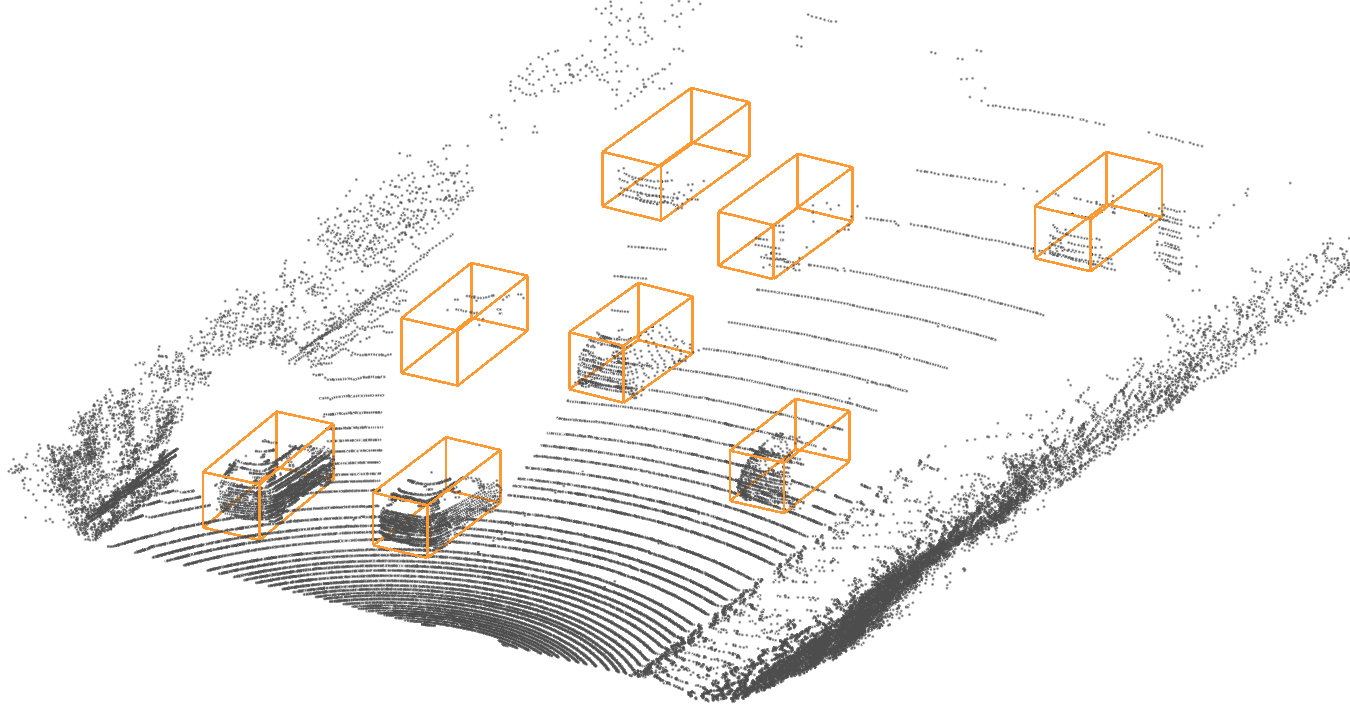}
    \hspace*{-0.9em}
    \includegraphics[width=0.5\linewidth]{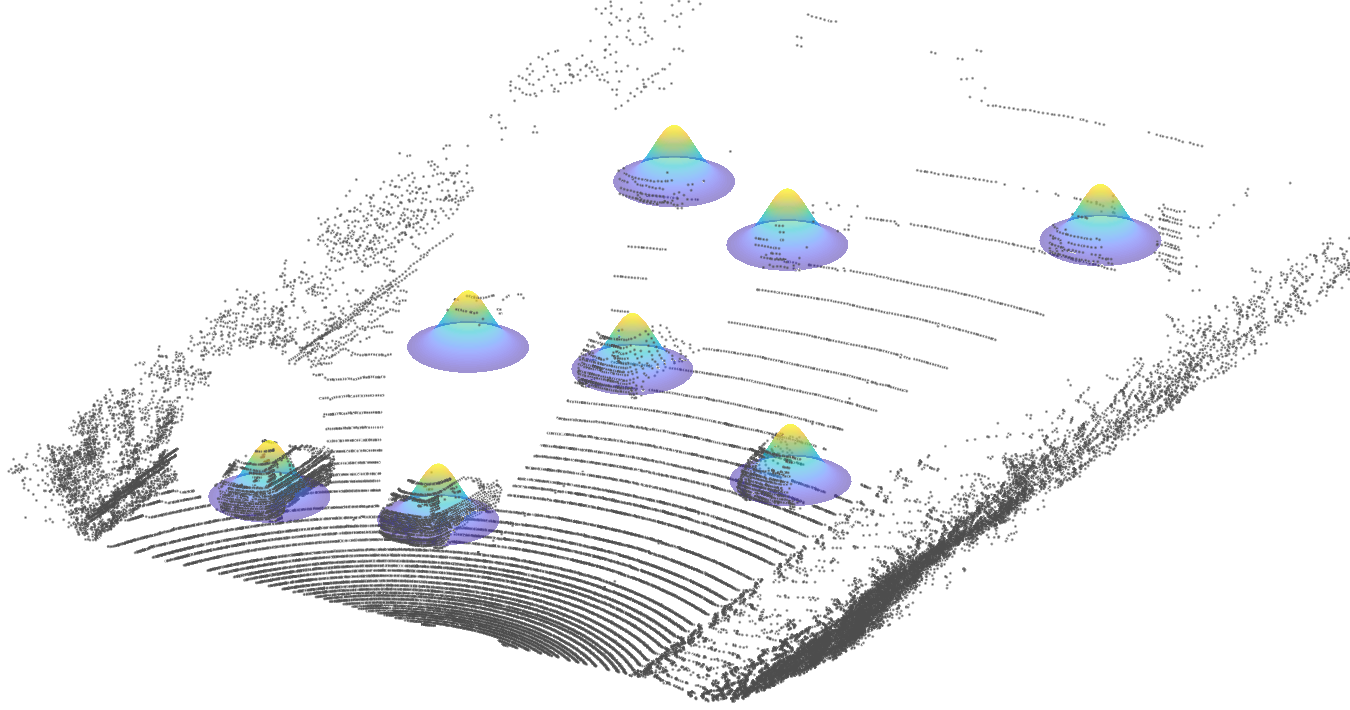}\\
    \vspace{0.25cm}
    \includegraphics[width=0.9\linewidth]{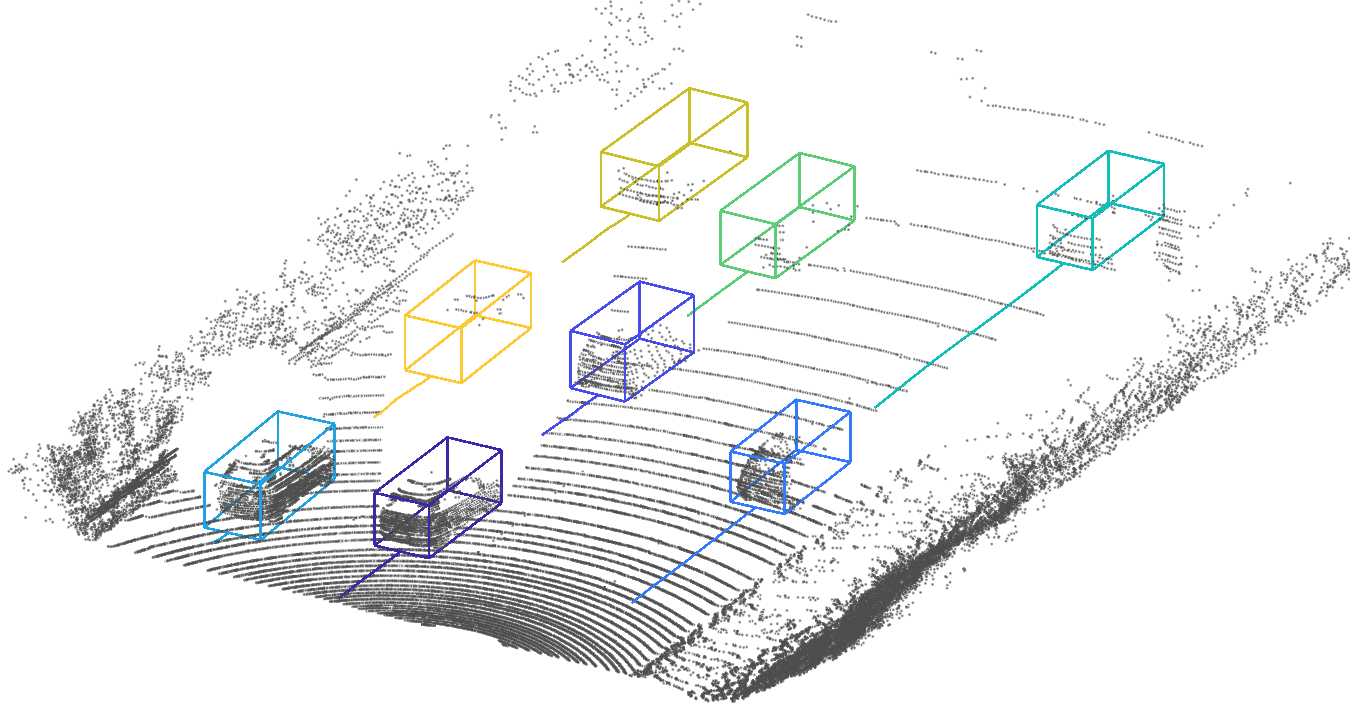}
    \caption{Visualization of the data flow in our algorithm. We use an off-the-shelf 3D object detector to obtain 3D bounding boxes (top left). In our algorithm, each detection is represented as a Gaussian distribution (top right).
    The outcome of our 3D multi-object tracker are optimized tracks in 3D space (bottom).}
    \label{fig:point clouds}
\end{figure}

Due to recent advances in 2D and 3D object detection \cite{FasterRCNN,FrustumPointNet,PointRCNN}, the majority of multi-object tracking algorithms follow the tracking-by-detection paradigm \cite{GlobalAssociation,Baseline,NMOT,BeyondPixels}.
The key challenge in this approach is the data association step, where new detections get assigned to existing tracks. 
Offline methods attempt to find the global optimal solution, for example through the use of min-cost flow algorithms \cite{GlobalAssociation} over the whole sequence.
Online tracking on the other hand must be fast and cannot wait for future detections.
Therefore, a common approach is to consider only detections of the current frame and assign them directly via the Hungarian algorithm \cite{SimpleOnline,Baseline}.

Since data association is the most crucial step for current tracking-by-detection algorithms, a common way is to formulate an assignment strategy that is as robust as possible.
In contrast to this approach, we propose a novel 3D multi-object tracking algorithm, which does not rely on explicit data association.
Instead, we propose a robust optimization back-end which is able to solve the assignment problem implicitly and jointly with the state estimation.
Based on the 3D bounding boxes from a neural network \cite{PointRCNN}, we create a Gaussian mixture model (GMM) to represent all detections simultaneously.
By using a factor graph formulation, we can apply the Max-Mixture \cite{Olson2012} to represent the full GMM in the state estimation process of each object.
In combination with inter-object constraints and a simple motion model, the full 3D position of each object can be estimated without solving the assignment problem explicitly.
This approach is motivated by the wide use of factor graphs in robotics for SLAM \cite{Dellaert2017} and general sensor fusion \cite{Pfeifer2019}.

Our main contributions are: \\
\textbf{Hybrid online/offline tracker:}
Our 3D multi-object tracker can be applied for offline as well as online tracking without any adaptions due to the flexible structure of factor graphs and their ability to keep track of the entire history of all states. 
We demonstrate in our experiments, that the offline solution achieves better results by considering all detections of a sequence.\\ 
\textbf{Implicit data association:} 
We do not need an explicit data association step since it is managed inside the factor graph as part of the optimization problem. 
Furthermore, this association is not fixed and can be changed during the optimization process, in contrast to most online trackers. \\
\textbf{Optimization of state positions:} 
We do not only estimate the association between states and detections, as min-cost flow approaches do, but also optimize the state positions over the whole sequence to deal with inaccuracies of the object detector.

In order to verify the ideas behind our novel approach and to demonstrate its capability for multi-object tracking, we conduct real-world experiments on the challenging KITTI MOT benchmark \cite{KITTI}. 
The proposed algorithm achieves accurate, robust and reliable tracking results and performs better than many state-of-the-art algorithms. 
Especially the consistency of the estimated tracks is superior in the offline as well as in the online case.
Considering the simplicity of our approach, these excellent results demonstrate the effectiveness of factor graph based 3D multi-object tracking.

% reset head height
\setlength{\headheight}{0pt}
\section{Related Work}
\subsection{3D Object Detection}
Reliable and accurate object detection is a crucial component of tracking algorithms, which follow the tracking-by-detection paradigm.
Prior work in the domain of 3D object detection can be roughly categorized into three classes. 
The algorithms proposed in \cite{Ku2019} and \cite{Xu2018} use only 2D images to directly predict 3D object proposals using neural networks. 
Another common approach is the combination of 2D images and 3D point clouds through neural networks to obtain 3D bounding boxes \cite{Ku2018,Chen2017}.
Methods of the third category solely rely on 3D point clouds and either project them to a 2D bird's-eye view \cite{Yang2018}, represent them as voxels \cite{VoxelNet} or directly extract 3D bounding boxes from raw point clouds \cite{PointRCNN}.  

\subsection{Multi-Object Tracking}
Following the common tracking-by-detection approach, offline trackers try to find the global optimal solution for the data association task over whole sequences. Typical methods are min-cost flow algorithms \cite{GlobalAssociation}, Markov Chain Monte Carlo \cite{Choi} or linear programming \cite{Berclaz}.
In contrast, online tracking algorithms only rely on past and current information and need to be real-time feasible. Common are filter-based approaches like Kalman \cite{Baseline} or particle filter \cite{Breitenstein}.
The data association step is often formulated as a bipartite graph matching problem and solved with the Hungarian algorithm \cite{Baseline,SimpleOnline}.
In the domain of 3D multi-object tracking a lot of work focuses on neural network based approaches, especially end-to-end learned models like \cite{Zhang,Luo}.
Both jointly learn 3D object detection and tracking.
The authors of \cite{Luo} rely solely on point clouds as input, while \cite{Zhang} utilizes camera images and point clouds.
Another approach is the incorporation of neural networks into filter-based solutions. 
The authors of \cite{Scheidegger} use 2D images as input for a deep neural network and combine it with a Poisson multi-Bernoulli mixture filter to obtain 3D tracks.

\subsection{Factor Graphs}
Although factor graphs are widely used in the field of robotics for \cite{Dellaert2017, Pfeifer2019}, they are not common in the tracking community. 
The authors of \cite{Schiegg} focus on solving the data association for 2D cell tracking, but do not optimize the cell positions. 
In \cite{Wang} the data association for multi-object tracking is solved using a factor graph in a 2D simulation, but an extended Kalman filter is applied for track filtering and prediction.
Therefore, the approach is not able to change the initial data association in past states.
Other work focuses on tracking with multiple sensors \cite{Meyer2017} or between multiple agents \cite{Meyer2016} and solves it via particle-based belief propagation in 2D space.
Their evaluation however, is limited to simulations and an application to real word use cases is unclear.

In contrast to prior work, we want to introduce online/offline capable, factor graph based multi-object tracking in 3D space. 
Furthermore, our approach is able to jointly describe data association and state positions for current and past states in a single factor graph and solve it effectively via non-linear least squares optimization.

\section{Multimodality and Factor Graphs}
\subsection{Factor Graphs for State Estimation}
\begin{figure}[tbph]
\centering
\includegraphics[width=0.9\linewidth]{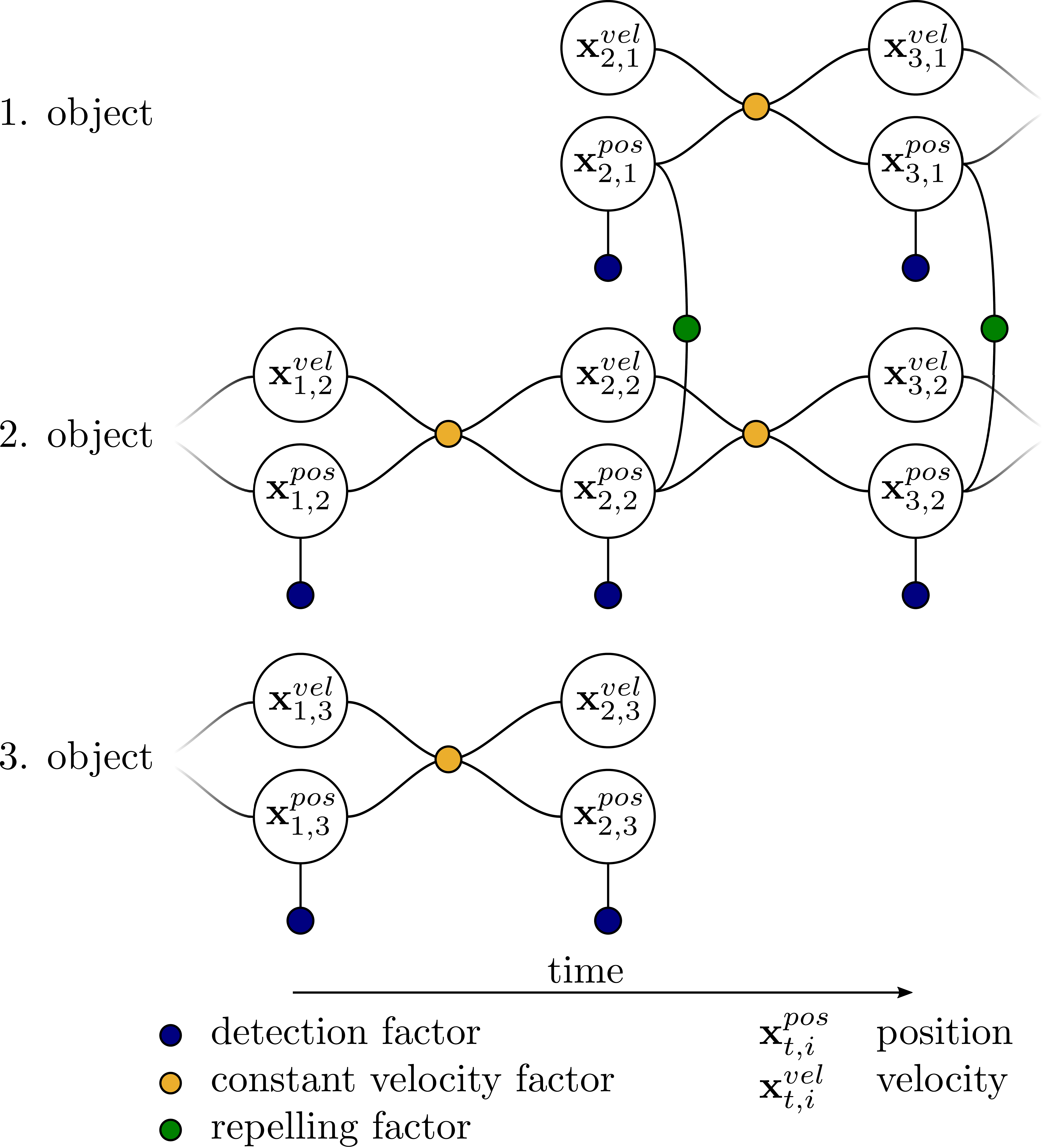}
\caption{Example of the factor graph representation of the proposed 3D multi-object tracking algorithm. Small dots represent error functions (factors) that define the least squares problem and big circles are the corresponding state variables. The set of tracked object varies over time due to the appearance and disappearance of objects from the field of view.}
\label{fig:FactorGraph}
\end{figure}

The factor graph, as a graphical representation of non-linear least squares optimization, is a powerful tool to solve complex state estimation problems and dominates today's progress in state estimation for autonomous systems \cite{Dellaert2017}.
\begin{equation} 
\StateOpt =\argmax_{\State} \prob(\State|\Measurement)
\label{eqn:argmax}
\end{equation}
To estimate the most likely set of states $\StateOpt$ based on a set of measurements $\Measurement$, \autoref{eqn:argmax} is solved using the factorized conditional probabilities:
\begin{equation}
    \probOf{\State}{\Measurement} \propto \prod_i \probOf{\measurementi}{\statei} \cdot \prod_j \prob(\statej)
    \label{eqn:posteriori}
\end{equation}
Please note that we omit the priors $\prob(\statej)$ in further equations for simplicity.
By assuming a Gaussian distributed conditional $\probOf{\measurementi}{\statei}$, the maximum-likelihood problem can be transformed to a minimization of the negative log-likelihood:
\begin{equation}
\StateEst = \argmin_{\State} \sum_{i} \frac{1}{2} \left\| \sqrtinfo \left( \errori - \mean \right) \right\| ^2
\label{eqn:arg_min}
\end{equation}
Therefore, the estimated set of states $\StateEst$ can be obtained by applying non-linear least squares optimization of the measurement function $\errori = f(\measurementi, \statei)$.
Additionally, the mean $\mean$ and square root information $\sqrtinfo$ of a Gaussian distribution are used to represent the sensor's error characteristic.
State-of-the-art frameworks like GTSAM \cite{Dellaert} or Ceres \cite{Agarwal} allow an efficient solution for online as well as offline estimation problems.

The major advantage over traditional filter-based solutions is the flexibility to re-estimate past states with the usage of current information.
This enables the algorithm to correct past estimation errors and improves even the estimation of current states.
Our algorithmic goal is the application of this capability to overcome the limitation of fixed assignments, that filters have to obey.

\begin{algorithm}[tbp]
	\SetAlgoLined
	generate detections $\Measurement$ using PointRCNN \cite{PointRCNN} 
	
	\ForEach{time step $t$}
	{
	    create GMM based on $\measurementtj$ and null-hypothesis
	    
	    \eIf{$t == 0$}
	    {
            init $\statePos$ at $\measurementPos$ and $\stateVel=[0,0,0]$ 
        }{
            propagate $\statePosPast$ to $t$
            
            get correspondence between $\stateti$ and $\measurementtj$ according to $-\log \left( \probOf{\measurementtj}{\stateti} \right)$
            
            \If{$\stateti$ does not correspond to any $\measurementtj$}
            {
                keep $\stateti$ marked as lost or delete it
            }
            
            \If{$\measurementtj$ does not correspond to any $\stateti$}
            {
                init $\statePos$ at $\measurementPos$ and $\stateVel=[0,0,0]$ 
            }
        }
        
        add factors \autoref{eqn:arg_min_mm}, \autoref{eqn:error_cv} and \autoref{eqn:error_rep}
            
        optimize factor graph
    }
    
    get association between $\State$ and $\Measurement$
    
    \eIf{$\stateti$ has associated $\measurementtj$}
    {
        $\stateDim = \measurementDim$ and $\stateConf = \measurementConf$
    }{
        $\stateDim = \stateDimPast$ and $\stateConf=0$
    }
    
    \If{$\meanOp\left( \stateConf \text{ } \forall \text{ } t \right) < \threshConf$} 
    {
        delete $\statei$
    }
    \caption{Tracking Algorithm for the Offline Case}
    \label{algo:tracking}
\end{algorithm}

\subsection{Solving the Assignment Problem}

Despite their capabilities, factor graphs are facing the same challenges as filters when it comes to unknown assignments between measurements and states.
The assignment can be represented by a categorical variable $\Params = \left\{ \ParamsSingle \right\}$ which describes the probability of the $j\text{th}$ detection  to belong to $i\text{th}$ object.
To solve the assignment problem exactly, the following integral over $\Params$ has to be solved:
\begin{equation}
    \probOf{\State}{\Measurement} = \int \probOf{\State, \Params}{\Measurement} \mathrm{d}\boldsymbol{\Params}
	\label{eqn:hidden}
\end{equation}
Common filter-based solutions estimate $\Params$ once, based on the predicted states, and assume it to be fixed in the following inference process.
This leads to a decreased performance in the case of wrong assignments.

Instead of including wrong assumptions, we formulate the state estimation problem without any assumptions regarding the assignment.
Therefore, we assume that $\Params$ follows a discrete uniform distribution.
This can be done by describing the whole set of measurements with an equally weighted Gaussian mixture model (GMM):
\begin{equation}
    \begin{split}
    \probOf{\measurementi}{\statei} \propto \sum_{j=1}^n&{c_j \cdot \exp \left( -\frac{1}{2} \left\| \sqrtinfoj \left( \errori - \mean_j \right) \right\| ^2 \right)}\\
    \text{with } &c_j = \weight_j \cdot \det\left( \sqrtinfoj \right)
    \end{split}
    \label{eqn:gmm}
\end{equation}
The GMM encodes that each measurement $\measurement_j$ with mean $\mean_j$ and uncertainty $\sqrtinfoj$ is assigned to each state $\statei$ with the same probability.
In our case, the error function is identical with the corresponding state ($\errori = \statei$).

By assigning all measurements to all states, the assignment problem has to be solved during inference by combining all available information.
It also allows to re-assign measurements and correct wrong matches with future evidence, which relaxes the requirement of an optimal initial assignment.

Using a GMM as probabilistic model breaks the least squares formulation derived in \autoref{eqn:arg_min}, which is limited to simple single Gaussian models.
The authors of \cite{Olson2012} proposed an approach to maintain this relationship by approximating the sum inside \autoref{eqn:gmm} by a maximum-operator.
This allows to reformulate the weighted error function as follows:
\begin{equation}
    \begin{split}
    \left\| \errorDet \right\|^2 
    &=
    \min_{j}
    \begin{Vmatrix}
    \sqrt{- 2 \cdot \ln \frac{c_j}{\gamma_m}}\\
    \sqrtinfoj \left( \errori - \mean_j \right)
    \end{Vmatrix}
    ^2\\
    \text{with } \gamma_m &= \max_{j} c_j
    \end{split}
    \label{eqn:arg_min_mm}
\end{equation}
For a detailed explanation of this equation, we refer the reader to our previous work \cite{Pfeifer2019} and the original publication \cite{Olson2012}.
Due to the flexibility of factor graphs, additional information like different sensors or motion models can be added to the optimization problem.

This combination of factor graphs and multimodal probabilistic models allows to formulate a novel inference algorithm as robust back-end for different multi-object tracking applications, that can be applied online as well as offline.

\section{Factor Graph based Tracking}
\begin{figure}[htbp]
    \centering
    \vspace{0.2cm}
    \includegraphics[width=0.6\linewidth]{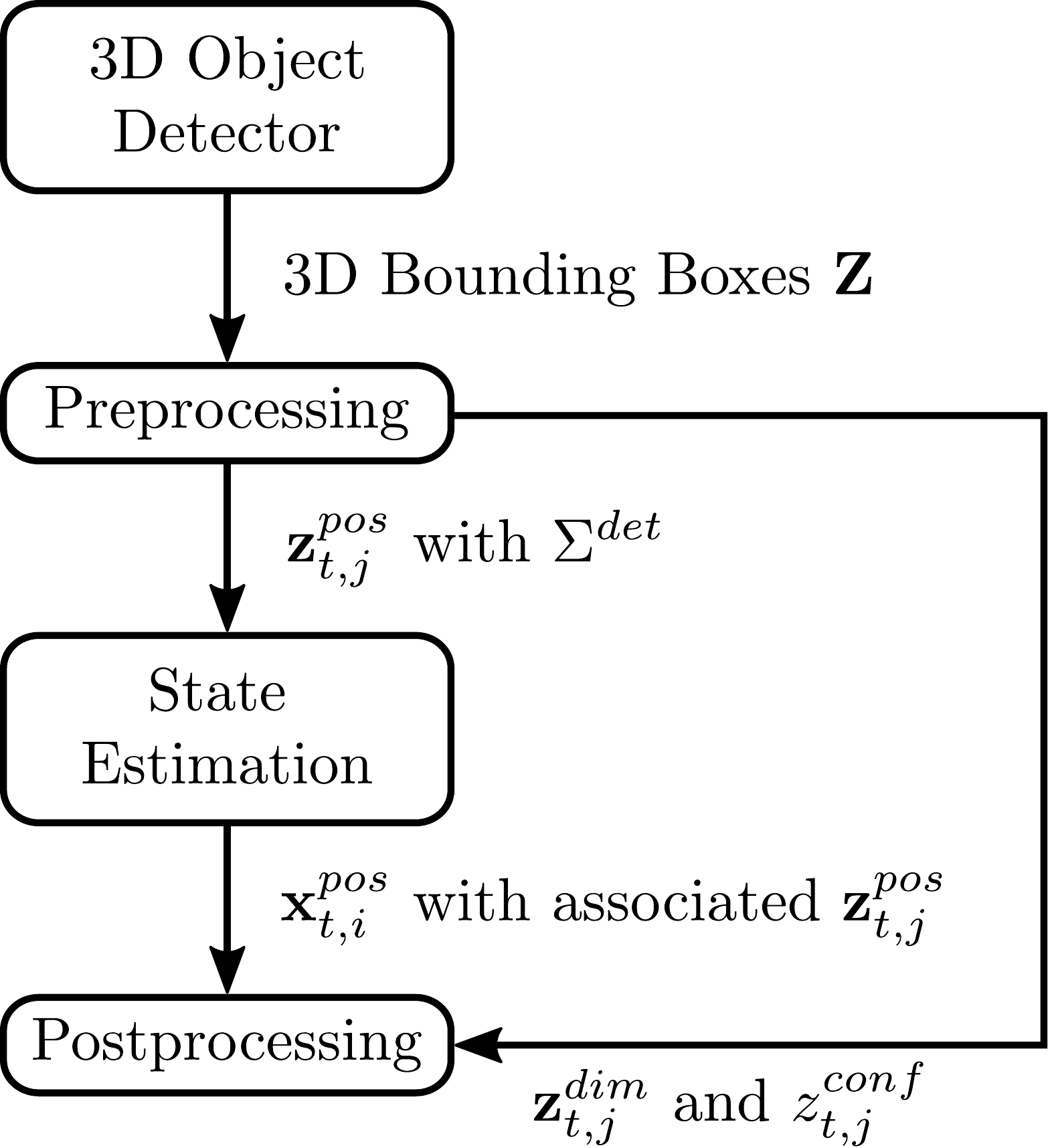}
    \caption{Block diagram of the introduced algorithm. The bounding box center points and their assignment to detections are optimized using a factor graph approach. The bounding box dimensions and confidence are adopted after the assignment.}
    \label{fig:overview}
\end{figure}

An overview of our approach to 3D multi-object tracking is given in \autoref{fig:overview} and the data flow is shown in \autoref{fig:point clouds}.
First we apply an off-the-shelf 3D object detector to obtain 3D bounding boxes.
Subsequently, all detections are represented as Gaussian distributions with the bounding box center as mean and $\covDet$ to describe the sensor's error characteristic.
Our algorithm jointly estimates the state positions in 3D space and solves the data association implicitly as part of the optimization.
In a postprocessing step the 3D bounding boxes are reconstructed through the combination of the states 3D positions and the association to the bounding boxes of the detector.
Furthermore, we filter out tracks with low confidence.
Details are explained in the following sections and the whole algorithm is shown in \autoref{algo:tracking} for the offline case and in \autoref{algo:tracking_online} for the online solution.

\subsection{Detection and Preprocessing}
We apply PointRCNN \cite{PointRCNN} to obtain the 3D bounding boxes $\Measurement$, which are composed of the 3D coordinates of the object's center, its dimensions, the rotation of the bounding box and its confidence and is defined at time step $t$ as:
\begin{equation}
\begin{split}
\measurementtj &= \begin{bmatrix}
            \measurementPos\\
            \measurementDim\\
            \measurementConf
            \end{bmatrix}\\
            \text{with }
\measurementPos = &\begin{bmatrix}
            z^x_{t,j}\\
            z^y_{t,j}\\
            z^z_{t,j}
            \end{bmatrix}
\measurementDim = \begin{bmatrix}
            z^h_{t,j}\\
            z^w_{t,j}\\
            z^l_{t,j}\\
            z^\theta_{t,j}
            \end{bmatrix} \\
\end{split}
\label{eqn:detections}
\end{equation}
Subsequently, all detections $\Measurement$ are transformed into a global coordinate system, which is defined relative to the ego vehicles pose at the first frame of the sequence.

\subsection{State Estimation}
The estimated states within the factor graph are composed of the 3D position $\statePos$ of the $i = 1 \dots M$ objects and their corresponding velocities $\stateVel$, both are defined at time step $t$ as:
\begin{equation}
\statePos = \begin{bmatrix}
            p^x_{t,i}\\
            p^y_{t,i}\\
            p^z_{t,i}
            \end{bmatrix}
\text{        }
\stateVel = \begin{bmatrix}
            v^x_{t,i}\\
            v^y_{t,i}\\
            v^z_{t,i}
            \end{bmatrix}
\label{eqn:states}
\end{equation}

\begin{algorithm}[tbp]
	\SetAlgoLined
	generate detections $\Measurement$ using PointRCNN \cite{PointRCNN} 
	
	\ForEach{time step $t$}
	{
	    create GMM based on $\measurementtj$ and null-hypothesis
	    
	    \eIf{$t == 0$}
	    {
            init $\statePos$ at $\measurementPos$ and $\stateVel=[0,0,0]$ 
        }{
            propagate $\statePosPast$ to $t$
            
            get correspondence between $\stateti$ and $\measurementtj$ according to $-\log \left( \probOf{\measurementtj}{\stateti} \right)$
            
            \If{$\stateti$ does not correspond to any $\measurementtj$}
            {
                keep $\stateti$ marked as lost or delete it
            }
            
            \If{$\measurementtj$ does not correspond to any $\stateti$}
            {
                init $\statePos$ at $\measurementPos$ and $\stateVel=[0,0,0]$ 
            }
        }
        
        add factors \autoref{eqn:arg_min_mm}, \autoref{eqn:error_cv} and \autoref{eqn:error_rep}
            
        optimize factor graph
        
         get association between $\stateti$ and $\measurementtj$
    
        \eIf{$\stateti$ has associated $\measurementtj$}
        {
            $\stateDim = \measurementDim$ and $\stateConf = \measurementConf$
        }{
            $\stateDim = \stateDimPast$ and $\stateConf=0$
        }
    
        \If{$\meanOp\left( \stateConf \text{ } \forall \text{ } t \right) < \threshConf$} 
        {
            do not output $\stateti$
        }
    }
    \caption{Tracking Algorithm for the Online Case}
    \label{algo:tracking_online}
\end{algorithm}

The detection factor is defined at each time step using \autoref{eqn:arg_min_mm} with the components mean $\mean_j = \measurementPos$ and a fixed uncertainty $\sqrtinfoj = (\covDet)^{-\frac{1}{2}}$ which corresponds to the detector's accuracy.
A generic null-hypothesis with a broad uncertainty $\covDetNull$ and mean $\meanDetNull = \meanOp\left( \measurementPos \text{ } \forall \text{ } j \right)$ is added to provide robustness against missing or wrong detections.
We use a simple constant velocity factor to describe the vehicle's motion:
\begin{equation}
    \left\| \errorCV \right\|^2_{\covCV} = 
    \left\|
        \begin{matrix}
        \left( \statePos - \statePosFut \right) - \stateVel \cdot \Delta t\\
        \stateVel - \stateVelFut
        \end{matrix}
    \right\|^2_{\covCV}
\label{eqn:error_cv}
\end{equation}
Please note, that $ \left\| \cdot \right\|^2_{\cov}$ denotes the Mahalanobis distance with the covariance matrix $\cov$.
To prevent two objects from occupying the same space, we add another simple constraint, which punishes the proximity of two objects. We call this factor the repelling factor:
\begin{equation}
    \left\| \errorRep \right\|^2_{\covRep} = 
    \left\|
        \frac{1}{\left\|\statePosn - \statePosm\right\|}
    \right\|^2_{\covRep}
\label{eqn:error_rep}
\end{equation}

The overall factor graph consists of one detection factor \autoref{eqn:arg_min_mm} per estimated object and corresponding constant velocity factors \autoref{eqn:error_cv} that connect the states of one object over time.
Repelling factors \autoref{eqn:error_rep} are added between object pairs if the euclidean distance is below a defined threshold $\threshDist$.
A visualization of the constructed factor graph is shown in \autoref{fig:FactorGraph}.

\begin{figure}[tpb]
    \centering
    \vspace{0.2cm}
    \includegraphics[width=0.5\linewidth]{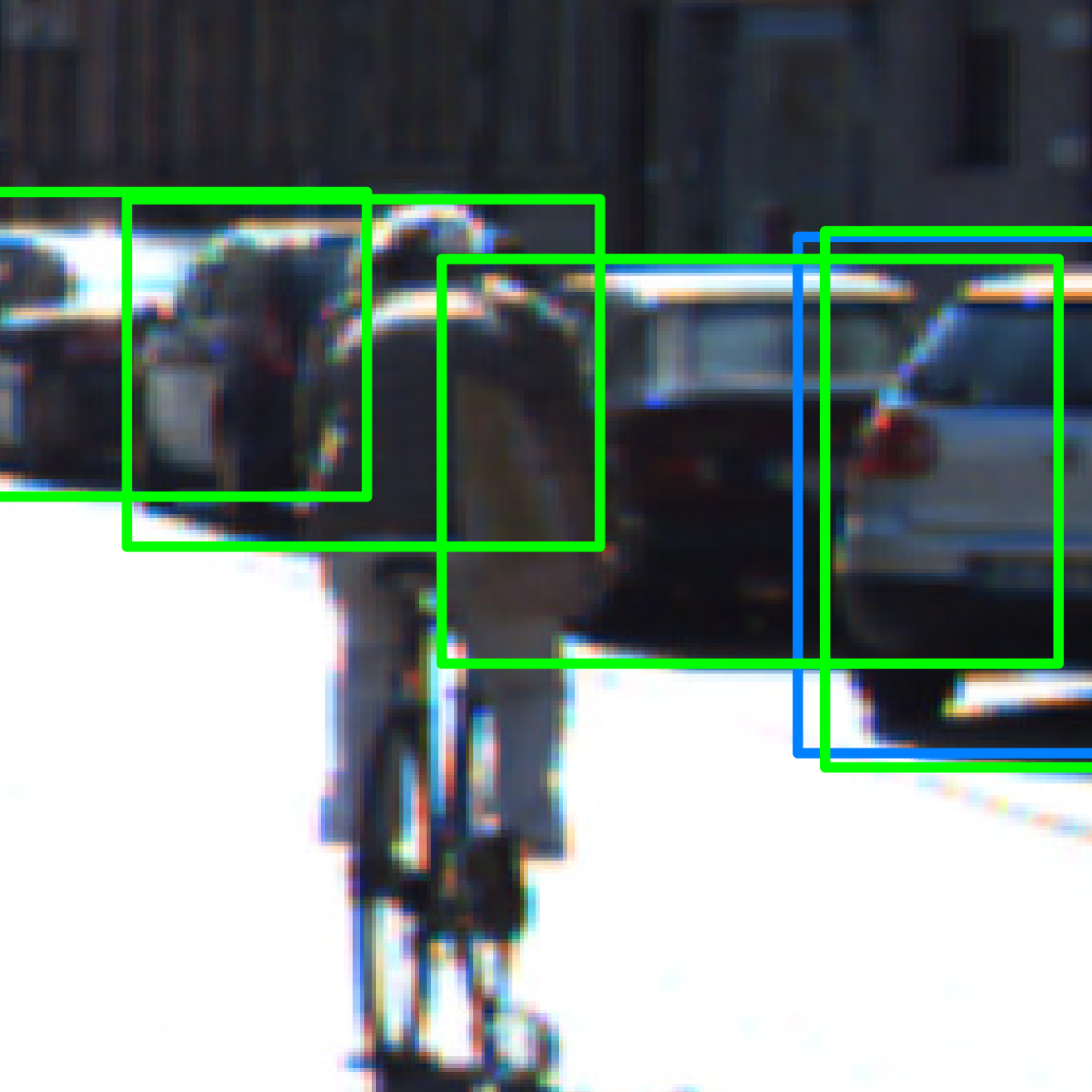}
    \hspace*{-0.7em}
    \includegraphics[width=0.5\linewidth]{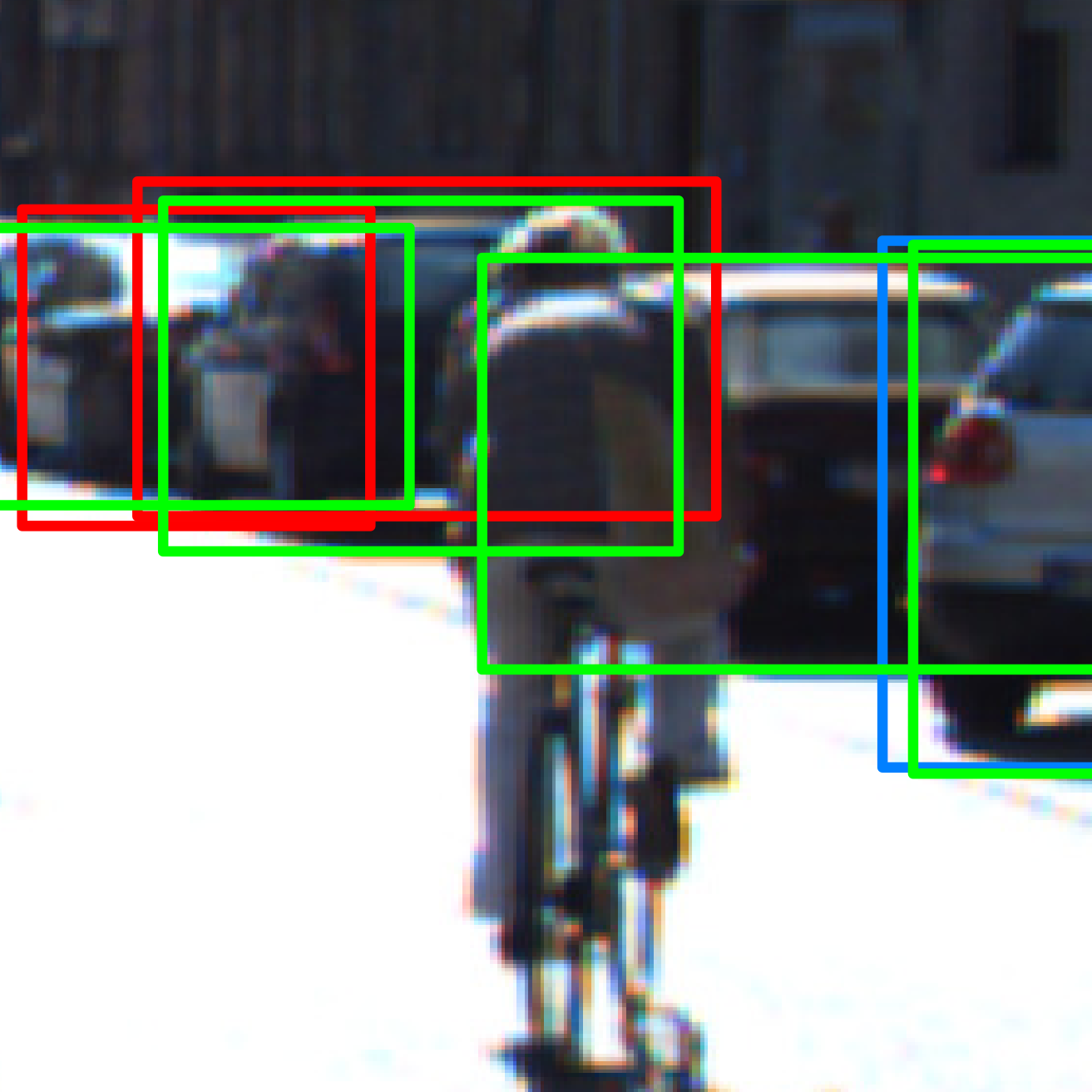}
    \caption{Limitations of the KITTI Ground Truth: Visualized are \textit{Don't Care} areas (red), \textit{Cars} (blue) and the results of our algorithm for class \textit{Car} (green). Due to the inconsistent labels we erroneously get 4 false positives (frames 109 and 110 from KITTI training sequence 0000).}
    \label{fig:kitti_gt}
\end{figure}

Our proposed algorithm is identical for offline and online tracking, except for the postprocessing step (compare \autoref{algo:tracking} and \autoref{algo:tracking_online}).
In the online use case it has to be done at each time step $t$. While it can be done once at the end for the offline solution.
At each time step $t$ new states can be added (creation of tracks) and existing states can be deleted (death of tracks) or carried over to the next time step (see \autoref{fig:FactorGraph}).
In order to create new tracks, we need to find detections $\measurementtj$ which do not correspond to any $\stateti$.
Therefore, we create a similarity matrix between all $\stateti$ (column) and $\measurementtj$ (row), including null-hypothesis $\measurementO$, by calculation of $-\log \left( \probOf{\measurementtj}{\stateti} \right)$.
Subsequently, we find the minimum (best similarity) and delete its state (column) from the matrix.
If the related measurement (row) is $\measurementO$ we mark the state as lost, since it does not correspond to any real measurement $\measurementtj$.
Otherwise, we also delete the measurement from the matrix, since it has a corresponding state.
These steps are repeated until all states are deleted from the matrix.
In that way, we can simultaneously find $\stateti$ which correspond to the null-hypothesis $\measurementO$ and $\measurementtj$ which do not correspond to any $\stateti$. 
A new $\stateti$ is created for each unrelated $\measurementtj$ in order to track it.
To suppress the tracking of false positive detections, all $\statei$ which have less consecutive detections than the defined threshold $\threshDet$ are deleted from the factor graph.
Thereby, the algorithm can simultaneously track objects from the first occurrence and suppress false positives in the offline use case.
This is not possible for the online solution.
Instead, states are only handed to the postprocessing step if they have $\threshDet$ or more consecutive detections.
In order to handle missing detections or occlusion of objects, a track $\statei$ has to be marked as lost (correspond to $\measurementO$) for more than $\threshLost$ consecutive time steps before it is terminated.
In this case, the last $\threshLost$ states $\stateti$ are deleted, since they do not correspond to real measurements. 
If a track has a corresponding $\measurementtj$ before hitting $\threshLost$, all $\statei$ are kept in the factor graph and the track lives on. 
Again, this is not possible in the online use case.
Instead, tracks $\statei$ marked as lost are only handed to the postprocessing step for a short duration of $\threshPerm$ time steps $t$. 

\subsection{Postprocessing}
After optimization, the factor graph returns the implicitly associated detection for each $\stateti$, which is either $\measurementtj$ or the null-hypothesis $\measurementO$.
Based on this, the optimized state positions $\statePos$ are combined with the matched bounding box $\stateDim = \measurementDim$ and confidence $\stateConf = \measurementConf$ or, in the case of the matched null-hypotheses, with the last bounding box of the same track $\stateDim = \stateDimPast$ and $\stateConf=0$.
Subsequently, we delete tracks $\statei$ with mean confidence below threshold $\threshConf$ for the offline solution (see \autoref{algo:tracking}).
In case of online tracking, the data association, bounding box fitting and track management has to be done at each time step $t$ (see \autoref{algo:tracking_online}). 
Therefore, we do not delete $\statei$ from the factor graph if the mean confidence is below threshold $\threshConf$, since the track can get above $\threshConf$ in the future. 
Instead, $\stateti$ is not handed to the postprocessing step and thus not part of the output. 

By filtering the online and offline results based on their confidence we can effectively throw away tracks with a high number of matched detections with low confidence, which are most likely false positives, and tracks which are matched primarily to the null-hypotheses. 
As a result, our algorithm is robust against false positives from the object detector.

\section{Experiments}
\subsection{Setup}
\begin{table}[tbp]
	\centering
	\vspace{0.25cm}
	\caption{Parameters of our FG-3DMOT algorithm}
	\label{tab:params_vbi}
	\resizebox{0.48\textwidth}{!}
    {%
	\begin{tabular}{@{}llc@{}}
		\toprule
		\textbf{Parameter Name}             & \textbf{Symbol}   &                 \textbf{Value}                                                                    \\ \midrule
		Detection Covariance                &  $\covDet$        &  $\diag \begin{pmatrix} \SI{0.2}{\meter}  \\ \SI{0.2}{\meter}\\ \SI{0.2}{\meter}\end{pmatrix}^2$  \\ \midrule
        Detection Null-Hypothesis Covariance&  $\covDetNull$    &  $\diag \begin{pmatrix} \SI{100}{\meter}  \\ \SI{100}{\meter}\\ \SI{1}{\meter}\end{pmatrix}^2$  \\ \midrule
		Constant Velocity Covariance        &  $\covCV$         &  $\diag \begin{pmatrix} \SI{0.25}{\meter}  \\ \SI{0.25}{\meter}\\ \SI{0.25}{\meter} \\ \SI{0.25}{\meter\per\second} \\ \SI{0.25}{\meter\per\second} \\ \SI{0.25}{\meter\per\second}\end{pmatrix}^2$ \\ \midrule
		Repelling Covariance                & $\covRep$         &  $\SI{0.5}{\meter^2}$ \\ \midrule
		Repelling Distance Threshold        & $\threshDist$     &  $\SI{4}{\meter}$  \\ \midrule
		Track Confidence Threshold (offline)& $\threshConf$     &  $3.9$  \\ \midrule
		Track Confidence Threshold (online) & $\threshConf$     &  $3.5$  \\ \midrule
		Number Consecutive Detections       & $\threshDet$      &  $2$  \\ \midrule
		Num. Con. Null-Hypothesis Detections   & $\threshLost$     &  $5$  \\ \midrule
		Object Permanence (online)          & $\threshPerm$     &  $1$ \\ \bottomrule
	\end{tabular}
	}
\end{table}

\begin{table*}[tbph]
    \centering
    \vspace{0.2cm}
    \caption{Results on the KITTI 2D MOT Testing Set for Class Car}
    \tiny
    \resizebox{0.83\textwidth}{!}
    {%
        \def\arraystretch{1.02}
        \begin{tabular}{@{} l  c  c  c  c  c  c  c @{}}
            \toprule
            {\bf Method} & {\bf MOTA $\uparrow$}  & {\bf MOTP $\uparrow$} & {\bf MT $\uparrow$} & {\bf ML $\downarrow$} & {\bf IDS $\downarrow$} & {\bf FRAG $\downarrow$} & {\bf FPS $\uparrow$}\\
            \midrule
            TuSimple \cite{NMOT} & 86.62 \% & 83.97 \% & 72.46 \% & 6.77 \% & 293 & 501 & 1.7 \\
            MASS \cite{MASS} & 85.04 \% & 85.53 \% & 74.31 \% & \bestResult{2.77} \% & 301 & 744 & 100.0 \\
            MOTSFusion \cite{MOTSFusion} & 84.83 \% & 85.21 \% & 73.08 \% & \bestResult{2.77} \% & 275 & 759 & 2.3  \\
            mmMOT \cite{Zhang} & 84.77 \% & 85.21 \% & 73.23 \% & \bestResult{2.77} \% & 284 & 753 & 50.0 \\
            mono3DT \cite{mono3DT} & 84.52 \% & 85.64 \% & 73.38 \% & \bestResult{2.77} \% & 377 & 847 & 33.3 \\
            MOTBeyondPixels \cite{BeyondPixels} & 84.24 \% & \bestResult{85.73} \% & 73.23 \% & \bestResult{2.77} \% & 468 & 944 & 3.3 \\
            AB3DMOT \cite{Baseline} & 83.84 \% & 85.24 \% & 66.92 \% & 11.38 \% & \bestResult{9} & 224 & \bestResult{212.8} \\
            IMMDP \cite{IMMDP} & 83.04 \% & 82.74 \% & 60.62 \% & 11.38 \% & 172 & 365 & 5.3 \\
            aUToTrack \cite{autotrack} & 82.25 \% & 80.52 \% & 72.62 \% & 3.54 \% & 1025 & 1402 & 100.0 \\
            JCSTD \cite{JCSTD} & 80.57 \% & 81.81 \% & 56.77 \% & 7.38 \% & 61 & 643 & 14.3 \\
            \midrule
            {\bf FG-3DMOT (offline)}  & \bestResult{88.01} \% & 85.04 \% & \bestResult{75.54} \% & 11.85 \% & 20 & \bestResult{117} & 23.8 \\ 
            {\bf FG-3DMOT (online)}   & 83.74 \% & 84.64 \% & 68.00 \% & 9.85 \% & \bestResult{9} & 375 & 27.1 \\
            \bottomrule
        \end{tabular}
    }
    \label{tab:result}
\end{table*}

In order to evaluate our proposed algorithm, we use the KITTI 2D MOT benchmark \cite{KITTI}. It is composed of 21 training and 29 testing sequences, with a total length of 8008 respectively 11095 frames. 
For each sequence, 3D point clouds, RGB images of the left and right camera and ego motion data are given at a rate of 10 FPS.
The testing split does not provide any annotations, since it is used for evaluation on the KITTI benchmark server.
The training split contains annotations for 30601 objects and 636 trajectories of 8 classes. 
Since annotations are rare for a lot of classes, we only evaluate our algorithm on the car subset.

As previously mentioned, we use PointRCNN \cite{PointRCNN} as 3D object detector.
Since \cite{Baseline} uses the same detector and made the obtained 3D bounding boxes publicly available, we use them for comparability reasons.
We normalize the provided confidence to positive values by adding a constant offset.
Since our algorithm is robust against false positive detections we achieve the best results by using all available detections.
All other parameters are shown in \autoref{tab:params_vbi}.

We construct the graph using the libRSF framework \cite{Pfeifer} and solve the optimization problem with Ceres \cite{Agarwal}, using the dogleg optimizer.

Since the KITTI 2D MOT benchmark is evaluated in the image space through 2D bounding boxes, we need to transform our 3D bounding boxes into the image space and flatten them into 2D bounding boxes. 
Furthermore, we only output 2D bounding boxes which overlap at least 25\% with the image in order to suppress detections that are not visible in the image, but in the laser scan.

\subsection{Results}\label{sec:results}

We evaluated our tracking algorithm in online and offline use case on the KITTI 2D MOT testing set.
Since offline results are rare and not among the best algorithms on the leaderboard, we compare our online and offline results against the 10 best methods on the leaderboard (accessed February 2020), which are summarized in \autoref{tab:result}.
The used metrics multi object tracking accuracy (MOTA), multi object tracking precision (MOTP), mostly tracked objects (MT), mostly lost objects (ML), id switches (IDS) and track fragmentation (FRAG) are defined in \cite{Li09learningto,Bernardin}.
The proposed algorithm achieves  accurate, robust and reliable tracking results in online as well as offline application, performing better than many state-of-the-art algorithms. 
In the offline use case, we improve the state-of-the art in accuracy, mostly tracked objects and the fragmentation of tracks.
Especially the fragmentation of our tracks is significantly lower than the previous state-of-the-art, since our algorithm can propagate tracks without any measurements far into the future and truncate them afterwards, if no measurements were associated.
Even for the online solution, we achieve  low track fragmentation and state-of-the-art id switches.
Furthermore, the results demonstrate that our approach benefits from the optimization of past states and the ability to re-assign past detections based on future information, since it achieves considerably higher accuracy and lower fragmentation in the offline application.
Although we did not optimize run time in any way, our algorithm is real-time feasible in online as well as offline use case. 
The computation time is similar in both cases and could be improved significantly, especially for the online use case. 
We provide a visualization of the results of our offline tracker in the accompanying video, which features KITTI testing sequence \textit{0014}\footnote[1]{https://youtu.be/mvZmli4jrZQ}. 

\textbf{Limitations of the KITTI Ground Truth:}
During the evaluation of our algorithm on the training split we discovered the limits of KITTI's 2D bounding box ground truth.
As an example we visualized a failure case of the ground truth in \autoref{fig:kitti_gt}.
The ground truth uses \textit{Don't Care} 2D bounding boxes to label regions with occluded, poorly visible or far away objects.
Bounding boxes which overlap at least 50\% with a \textit{Don't Care} area are not evaluated.
As seen in \autoref{fig:kitti_gt}, the ground truth is not labeled consistently and objects (here cars) are neither labeled \textit{Don't Care} nor \textit{Car}.
Therefore, we get a lot of false positives, since our algorithm can track objects from the first occurrence and from high distances.
As a result, we assume that we could achieve significantly higher accuracy (MOTA) with a consistent labeled ground truth.

\textbf{Limitations of our Algorithm:}
After the evaluation on the testing set of the KITTI benchmark we discovered one main failure case of our algorithm. 
In testing sequences \textit{0005} and \textit{0006} the ego-vehicle captures a highway scene, hence it self and all other cars are moving with high velocities.
Since our algorithm works in a static 3D coordinate system and we initialize new objects as a general assumption with zero velocity, these cases are hard to capture for our algorithm.
Furthermore, such scenes with fast moving cars are not present in the training split, whereby we could not adopt our algorithm to handle these scenes.
This issue will be addressed in future work.

\section{Conclusion}
Our proposed 3D multi-object tracking algorithm is build around three main ideas: Implicit and variable data association using the factor graph formulation, optimization of object positions in 3D space and the application as online and offline solution.
In order to prove the capabilities of our novel approach, we conducted real world experiments and achieved state-of-the-art results with a superior tracking consistency.
Based on this minimalist approach a vast amount of optimization possibilities are given.
Due to the flexibility of factor graphs, improvements like more sophisticated motion models or prior knowledge can be easily integrated into the current algorithm.
The integration of additional sensors, like cameras with own detection pipelines, could also be a direction of future research.
We will continue this work in the future, to exploit the capabilities of our multimodal measurement model for the even more difficult problem of extended object tracking.
The ability of factor graphs to store the complete history reveals a great potential in this field.
\vspace{0.2cm}

\bibliographystyle{IEEEtran}
\balance
\bibliography{IEEEabrv,references.bib}
~ ~ ~ ~ ~ ~ ~ ~ ~ ~\\
~ ~ ~ ~ ~ ~ ~ ~ ~ ~\\
\end{document}